\documentclass[10pt,twocolumn,letterpaper]{article}

\usepackage[pagenumbers]{cvpr} % To force page numbers, e.g. for an arXiv version
\usepackage{adjustbox}
\usepackage{booktabs}
\usepackage{times}
\usepackage{epsfig}
\usepackage{graphicx}
\usepackage{amsmath}
\usepackage{amssymb}
\usepackage{multirow}
\usepackage{subcaption}
\usepackage[table,xcdraw,dvipsnames]{xcolor}
\usepackage{pifont}% http://ctan.org/pkg/pifont
\usepackage{algorithmic}
\usepackage{pifont}
\newcommand{\xmark}{\ding{55}}
\usepackage[linesnumbered,ruled,vlined]{algorithm2e}

\SetKwComment{Comment}{\color{green!50!black}\# }{}

\SetKwProg{Function}{def}{:}{}

\SetKwProg{For}{for}{:}{}
\SetKwProg{If}{if}{:}{}
\newcommand{\rowA}{\rowcolor{gray!15}}

\usepackage[pagebackref,breaklinks,colorlinks]{hyperref}

\usepackage[capitalize]{cleveref}
\crefname{section}{Sec.}{Secs.}
\Crefname{section}{Section}{Sections}
\Crefname{table}{Table}{Tables}
\crefname{table}{Tab.}{Tabs.}

\def\cvprPaperID{15147} % *** Enter the CVPR Paper ID here
\def\confName{CVPR}
\def\confYear{2024}

\newcommand{\ourmodel}{\textit{DINOv}}
\newcommand{\cmark}{\checkmark}%
\title{Visual In-Context Prompting}
\author{
  ~Feng Li$^{\spadesuit}$, ~~Qing Jiang$^{\sharp}$, ~~Hao Zhang$^{\spadesuit}$, ~~Tianhe Ren$^{\dagger}$, ~~Shilong Liu$^{\P}$, ~~Xueyan Zou$^{\S}$,  ~~Huaizhe Xu$^{\spadesuit}$, \\ ~~Hongyang Li$^{\sharp}$,~~Chunyuan Li$^{\ddagger}$,~~Jianwei Yang$^{\ddagger1}$,~~Lei Zhang$^{\dagger2}$,~~Jianfeng Gao$^{\ddagger2}$
\and
{
\footnotesize
$^{\spadesuit}$ HKUST \;
$^\ddagger$ Microsoft Research, Redmond \;  
$^\dagger$ IDEA \;  
$^{\sharp}$ SCUT \;
$^{\P}$ Tsinghua \;
$^{\S}$ UW-Madison \;
}
\and
\scriptsize{
$1.$~Project Lead \;
$2.$~Equal Advisory Contribution \;
}
}
\begin{document}
\twocolumn[{%
\renewcommand\twocolumn[1][]{#1}%
\maketitle
\vspace{-2.3em}
\begin{center}
    \centering
    \includegraphics[width=0.98\textwidth]{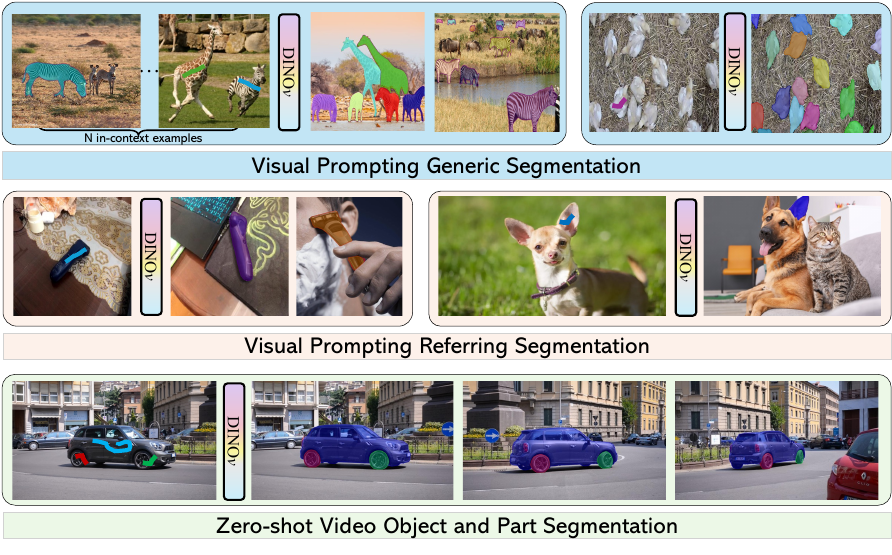}
% \vspace{0.2em}
\vspace{-1.3em}
\captionof{figure}{Our model \ourmodel{} supports generic and referring segmentation to associate multiple or single objects with the user input visual prompts. A user can input one or more in-context visual prompts (scribbles, masks, boxes, etc.) to improve the segmentation performance.
% Our model \ourmodel{} supports generic and referring segmentation. Generic: segment all objects of the same semantic concept that match the user prompt. Refer: segment a particular object with the user input visual prompts. (single) Visual prompt: one image-prompt example to segment. In-context prompt: one or multiple image-prompt examples. \ourmodel{} supports one or more in-context visual prompts (scribbles, masks, boxes, etc.) to improve the segmentation performance.
}
\label{fig:intro}
\vspace{0.5em}
\end{center}%
}]
% \maketitle

%%%%%%%%% ABSTRACT
\begin{abstract}
\vspace{-2mm}
    In-context prompting in large language models (LLMs) has become a prevalent approach to improve zero-shot capabilities, but this idea is less explored in the vision domain. Existing visual prompting methods focus on referring segmentation to segment the most relevant object, falling short of addressing many generic vision tasks like open-set segmentation and detection. In this paper, we introduce a universal visual in-context prompting framework for both tasks, as shown in Fig.~\ref{fig:intro}. In particular, we build on top of an encoder-decoder architecture, and develop a versatile prompt encoder to support a variety of prompts like strokes, boxes, and points. We further enhance it to take an arbitrary number of reference image segments as the context. Our extensive explorations show that the proposed visual in-context prompting elicits extraordinary referring and generic segmentation capabilities to refer and detect, yielding competitive performance to close-set in-domain datasets and showing promising results on many open-set segmentation datasets. By joint training on COCO and SA-1B, \ourmodel{} achieves $57.7$ PQ on COCO and $23.2$ PQ on ADE20K. Code will be available at \url{https://github.com/UX-Decoder/DINOv}
    % \jianwei{we may highlight some quantitative numbers here}0---------------------------------------------------
   % % In this paper, we introduce visual in-context prompting for fine-grained vision tasks.
   % This paper pioneers the application of visual in-context prompting to fine-grained vision tasks, introducing a unified model that adeptly handles both referring and generic image segmentation challenges. 
   % % We define the segmentation of a single instance as a one-to-one matching problem and the segmentation of multiple instances within a category as a one-to-many matching problem. 
   % Notably, our work is the first to effectively address the generic segmentation problem with visual prompts, thereby extending the capabilities of visual in-context learning to encompass general object detection and segmentation tasks.
   % The model is underpinned by three innovations: 1. A content embedder designed to extract region-level features for prompts, enabling the model to interpret visual interactions like strokes, boxes, and points, which are essential for one-to-many tasks. 2. The use of auto-generated SAM data for richer semantics in one-to-one tasks, expanding the model's instance vocabulary beyond traditional methods. 3. An elegant unified strategy to solve both one-to-one and one-to-many matching problems, employing transposed matching matrices and combined data for enhanced performance across tasks.
\end{abstract}

%%%%%%%%% BODY TEXT
\section{Introduction}
\label{sec:intro}

% why we need in context prompt?
The recent progress in large language models (LLMs) like GPT~\cite{brown2020language,openai2023gpt4} has shown promising results towards artificial general intelligence (AGI) by training unified models on large amounts of text data. These giant LLMs manifest themselves with intriguing emerging capabilities such as in-context learning. Nevertheless, similar paradigms have not yet succeeded in solving all vision tasks due to the diversity of scenarios in computer vision~\cite{li2023multimodal}. Some works~\cite{zhu2023minigpt4,liu2023llava} have combined LLMs and vision models to tackle complex image understanding tasks with text outputs such as visual question answering, but challenges remain in fine-grained tasks that require pixel-level outputs, like instance masks, rather than just text.

% Another approach to building a unified vision model is to incorporate LLM capabilities into vision architectures.
To this end, the community has observed a growing interest in the development of language-enhanced vision foundation models. These models demonstrate profound competencies in open-world visual understanding tasks using text prompts, encompassing areas like open-set detection~\cite{liu2023grounding, zhang2023simple, li2022grounded} and segmentation~\cite{zou2022generalized, zhang2023simple, yu2023convolutions, xu2023openvocabulary}. Visual prompt, a different prompting mechanism has been explored in some recent segmentation models~\cite{kirillov2023segment,zou2023segment, li2023semantic}. In these works, different visual prompting formats (\textit{e.g.}, points, boxes and strokes, \textit{etc}) have been explored to facilitate the segmentation of visual contents specified by users.

\begin{figure}[t!]
\centering  
\vspace{-0mm}
\includegraphics[width=1.00\linewidth]{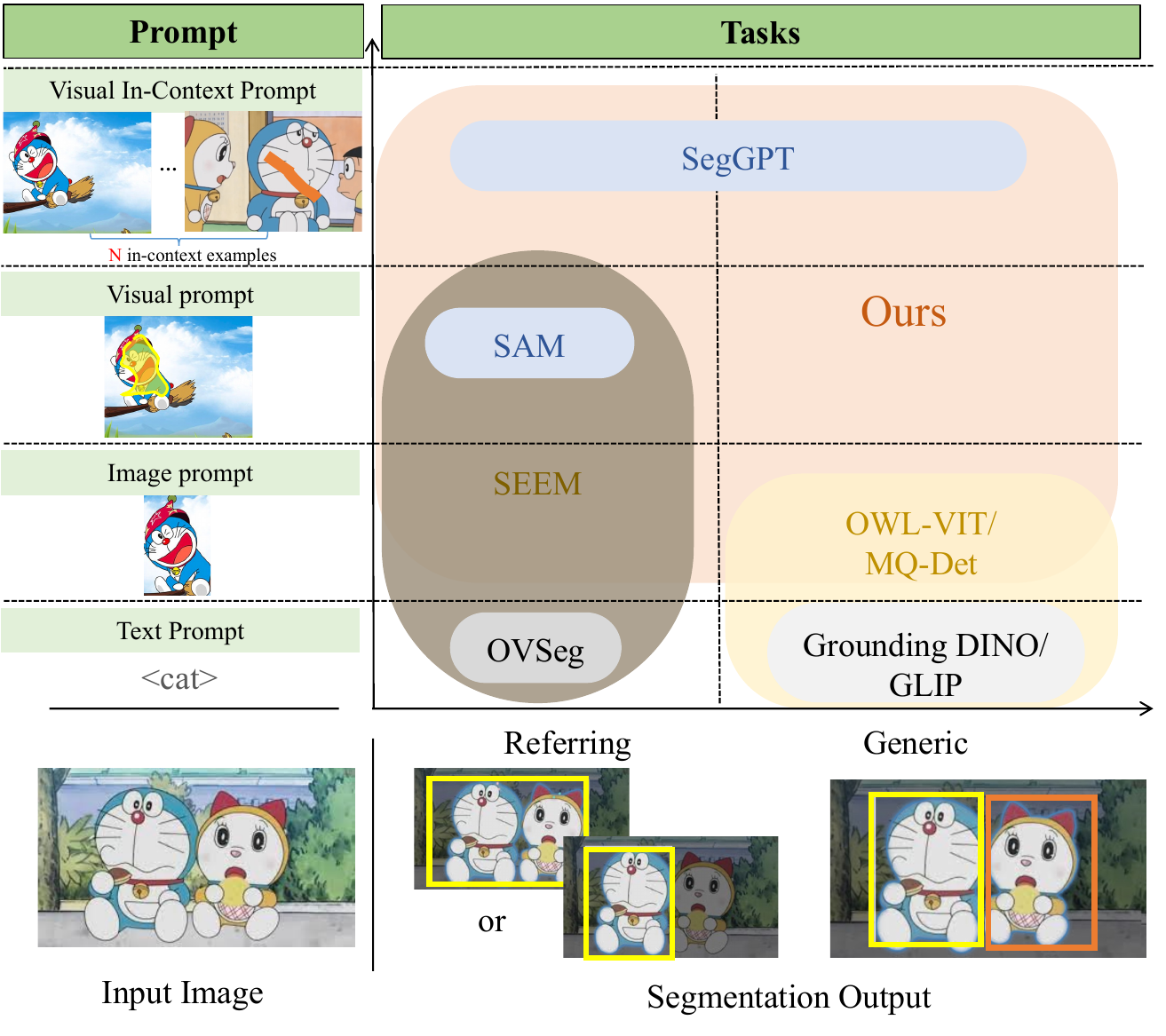} \\
\vspace{-2mm}
\caption{Comparison with related works. Generic: segment all objects of the same semantic concept that match the user prompt. Refer: segment a particular object with the user input visual prompts. Image prompt: crop the image regions as prompts. (single) Visual prompt: one image-prompt example to segment. In-context prompt: one or multiple image-prompt examples. We can do single-image and cross-image visual prompting tasks and support referring and generic segmentation.}
\label{fig:scope}  
  \vspace{-3mm}
\end{figure}

In-context learning, an appealing capability in LLMs, has been less explored. It specifies the new task instruction using examples, and allows models to adapt to new tasks or domains without explicit retraining. 
% by providing relevant examples.
% Previous studies have explored in-context learning in the context of image segmentation, where they focus on associating a user visual prompt with one most relevant objects. These approaches fall short of addressing standard object detection and segmentation tasks. 
One pioneering work for visual in-context learning is SegGPT~\cite{wang2023seggpt}, which demonstrates the ability to output an image mask based on visual examples. However, these works focus on associating a user visual prompt with one most relevant object and have the limited ability to identify multiple objects of the same semantic concept. More importantly, prompting in the image pixels with colorful masks inherently fails to generalize to novel concepts. As such, it is not competent to address many generic vision tasks like open-set object detection and segmentation, which often require the segmentation of multiple objects of a given concept. 
% but have limited ability to identify multiple objects of the same semantic concept.
 % within an image. This approach is inadequate for mainstream vision tasks, which often require the segmentation of multiple objects of a given concept. 
% Despite the above efforts to design visual prompting and visual in-context prompting models for segmentation tasks, we are still lacking a generic visual prompting model that can accommodate general vision tasks including open-set segmentation and detection. 
On the other hand, textual prompting in vision models exhibit notable flexibility in managing both referring and generic tasks in detection or segmentation~\cite{liu2023grounding,zou2022generalized}. Nevertheless, they are arguably not favorable for in-context settings in that they cannot take segmentation masks as the inputs. In this paper, we strive to develop a model that supports \emph{visual in-context prompting} for all types of image segmentation tasks. A comparison between our work and previous work is shown in Fig.~\ref{fig:scope}. Besides supporting both single-image and cross-image visual prompting, our model distinguishes itself by effectively handling both referring and generic segmentation problems.
% , a first in the field, achieving performance comparable to closed-set vision models on standard benchmarks. 

\begin{figure*}[t!]
    \centering

    \includegraphics[width=\textwidth]{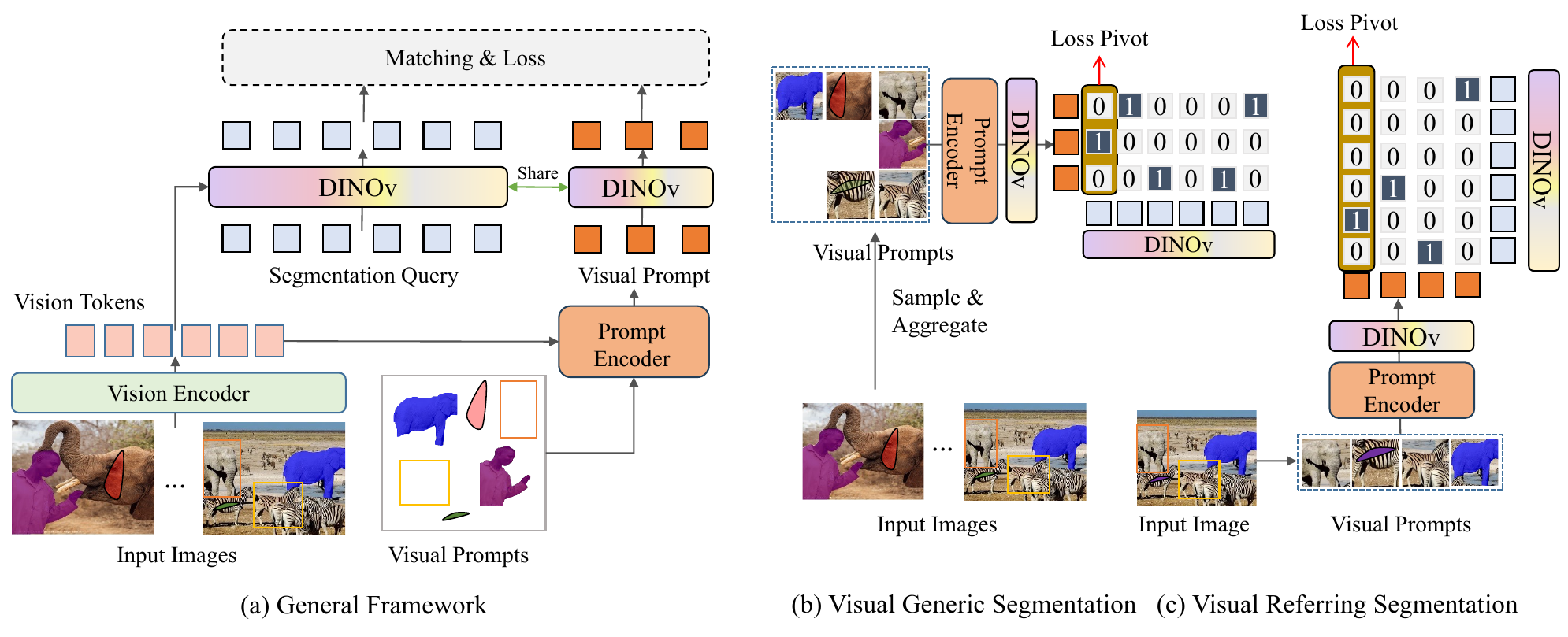}
    \vspace{-15pt}
    \caption{\ourmodel{} is a universal segmentation framework that can do generic segmentation and referring image segmentation. The vision encoder is used to extract image features.(b) An illustration of losses for visual generic segmentation. In the example, there are 6 visual prompts sampled from 6 masks from 3 categories. The visual prompts from the instances of the same class are averaged as the class embedding. Each colume of the matching matrix is a 3-dimension one-hot vector which is a one-hot class label of the instance; (c) An illustration of losses for visual referring segmentation. Each visual prompt is classified to one of the 6 instances.
    % (c) An illustration of losses for visual generic segmentation and visual referring segmentation. 
    }    
    % \hao{The caption is too short. Should explain each sub-figure. For (c), I understand each collume denotes a visual prompt, but the meaning of each row is not very clear.}
    \label{fig:framework_all}
    % \vspace{-12pt}
    \vspace{-5pt}
\end{figure*}

To achieve this goal, we build a model called \textbf{\ourmodel{}} to support versatile visual prompting capabilities, based on the unified detection and segmentation model MaskDINO~\cite{li2023mask}. \ourmodel{} follows the general encoder-decoder design with an extra prompt encoder to formulate and sample visual prompts. The decoder takes in segmentation queries and reference prompt queries to generate segmentation masks and target visual prompts, and we associate the output segmentation masks with the target prompt queries for the final output.  We can define the visual in-context samples with a set of \textit{reference image~(Q) - visual prompt~(A)} pairs. The visual prompt can be in various types, including mask, scribble, box, etc. With the in-context examples, our model takes in a target image and outputs the masks. The creation of target visual prompts involves an initial step where a prompt encoder extracts reference visual prompts from a Q-A pair. This is followed by a decoder to get the target visual prompt by attending reference visual prompts to the target image. During training, to construct positive and negative samples for generic segmentation, we sample reference visual prompts in a batch across different images. To address task and data discrepancies, we formulate generic latent queries and point queries for generic and referring segmentation, respectively. By joint training on COCO~\cite{lin2014microsoft} and SA-1B \cite{kirillov2023segment} for generic and referring segmentation, our model attains competitive performance on in-domain segmentation tasks compared with text-prompted models and shows promising generalization capability on a wide range of open-set segmentation benchmarks using purely visual prompts.

To summarize, our contributions are threefold: 
% \\\nointent\textbf{1}. We are the first work to extend visual in-context prompting to general vision tasks like open-set generic segmentation and detection tasks, and achieve comparable performance with text prompt-based open-set models.
% \\\nointent\textbf{2}. We build \ourmodel{}, a unified framework for referring segmentation and generic segmentation based on visual in-context prompting. This unification simplifies model design and allows our model to consume both semantically-labelled and unlabelled data for better performance.
% \\\nointent\textbf{3}. We conduct extensive experiments and visualizations to show that our model can handle generic, referring, and video object segmentation tasks.  Our early attempts exhibit promising results on open-set segmentation and detection with visual prompting.
\begin{itemize}
\vspace{-0.6em}
    \item We are the first to extend visual in-context prompting to support generic vision tasks like open-set generic segmentation and detection, and achieve comparable performance with text prompt-based open-set models.
    \vspace{-0.6em}
    \item We build \ourmodel{}, a unified framework for referring segmentation and generic segmentation based on visual in-context prompting. This unification simplifies model design and allows our model to consume both semantically-labelled and unlabelled data for better performance.
    \vspace{-0.6em}
    \item We conduct extensive experiments and visualizations to show that our model can handle generic, referring, and video object segmentation tasks.  Our early attempts exhibit promising results on open-set segmentation and detection with visual prompting.
    \vspace{-0.6em}
\end{itemize}

\section{Method}
\subsection{Unified Formulation for Segmentation Tasks}
In this paper, we concentrate on visual prompting tasks involving images, encompassing both generic segmentation and referring segmentation tasks.  Given $N$ reference images $\mathcal{I}=\{\mathbf{I}_1, ..., \mathbf{I}_N\} \in \mathcal{R}^{N \times H \times W \times 3}$ with the corresponding visual prompts $\mathcal{P}=\{p_1,...,p_N\}$, \ourmodel{} aims to segment objects of interest on a new target image $\mathbf{I_t}$. The visual prompts include masks, boxes, scribbles, points, etc. The interested objects can be a particular object for referring segmentation or all objects of the same semantic concept for generic segmentation. Note that the reference image can be identical to the target image, in which the task reduces to single-image visual prompting segmentation.
 % ~\cite{kirillov2023segment, zou2023segment, li2023semantic}.
\begin{figure*}[t!]
    \centering
    \includegraphics[width=0.98\textwidth]{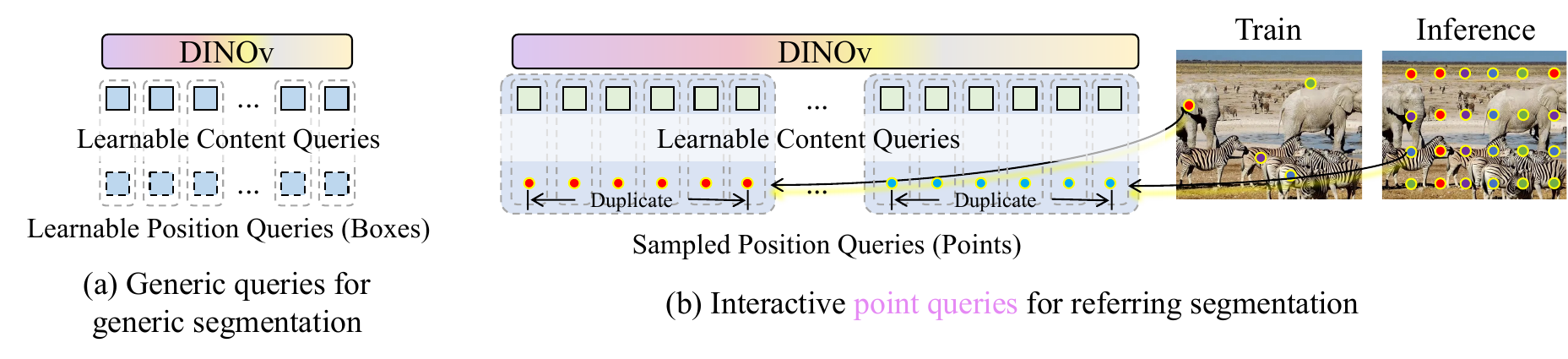}
    \vspace{-5pt}
    \caption{\ourmodel{} query formulation of generic and referring segmentation tasks.  }    
    \vspace{-5pt}
    \label{fig:query_formulation}
    % \vspace{-12pt}
\end{figure*}

To address these tasks, \ourmodel{} utilizes a comprehensive query-based encoder-decoder architecture. This architecture comprises a vision encoder, denoted as $\mathbf{Enc}$, responsible for extracting image features, a prompt encoder referred to as $\mathbf{PromptEncoder}$, designed to extract visual prompt features by combining image features and user-provided visual prompts, and a general decoder represented as $\mathbf{Decoder}$, which generates masks and visual concepts based on the segmentation query and visual prompt features. Upon receiving the input image and user-provided visual prompts, our initial step involves extracting image features denoted as $\mathbf{Z}$ using the vision encoder. Subsequently, we feed both the image features and visual prompts into the prompt encoder to extract the \textit{reference visual prompt} $\mathcal{F}$ and subsequently sample the \textit{query visual prompt} features $\mathbf{Q_p}$.Formally, we have:
 \begin{equation}
 \begin{aligned}
    \mathcal{Z}&=\mathbf{Enc}(\mathcal{I}), \mathbf{Z}=\mathbf{Enc}(\mathbf{I_t}) \\
    \mathcal{F}&=\mathbf{PromptEncoder}(\mathcal{P}, \mathcal{Z}) \\
    \mathbf{Q_p}&=\mathbf{PromptSample}(\mathcal{F}) \\
\end{aligned}
\label{EQ:encoder}
\end{equation}

In addition to the visual prompt features $\mathbf{Q_p}$, \ourmodel{} incorporates segmentation queries $\mathbf{Q_s}$ for proposal extraction. A shared decoder is employed to decode outputs for both $\mathbf{Q_s}$ and $\mathbf{Q_p}$ while performing cross-attention with respect to the target image feature $\mathbf{Z}$.

\begin{equation}
\begin{aligned}
    \mathbf{O_s} &= \mathbf{Decoder}\left(\mathbf{Q_s}; \mathbf{Z}\right) \\ \mathbf{O_p} &= \mathbf{Decoder}\left(\mathbf{Q_p}; \mathbf{Z}\right) \\
    \langle\mathbf{M}, \mathbf{B}\rangle&=\mathbf{MaskHead}(\mathbf{O_s}) \\
    \mathbf{C_{g},C_{r}}&=\mathbf{PromptClassifier}(\mathbf{O_s}, \mathbf{O_p}) \\
\end{aligned}
\label{EQ:decoder}
\end{equation}
% \hao{$\mathbf{C}$ seems to be one matrix, I prefer to use $\mathbf{C_{g},C_{r}}$, coherent with the next paragraph.}
Here, $\mathbf{O_s}$ represents the decoded segmentation query features, $\mathbf{O_p}$ corresponds to the decoded \textit{target visual prompt} features, while $\mathbf{M}$ and $\mathbf{B}$ denote the predicted masks and boxes, respectively. Furthermore, we have $\mathbf{C_{g}}$ and $\mathbf{C_{r}}$ as the predicted matching scores for generic segmentation and referring segmentation tasks. These scores are derived through the use of a $\mathsf{PromptClassifier}$, which computes the similarity between $\mathbf{O_s}$ and $\mathbf{O_p}$.

\begin{figure}[t!]
    \centering
    \includegraphics[width=0.42\textwidth]{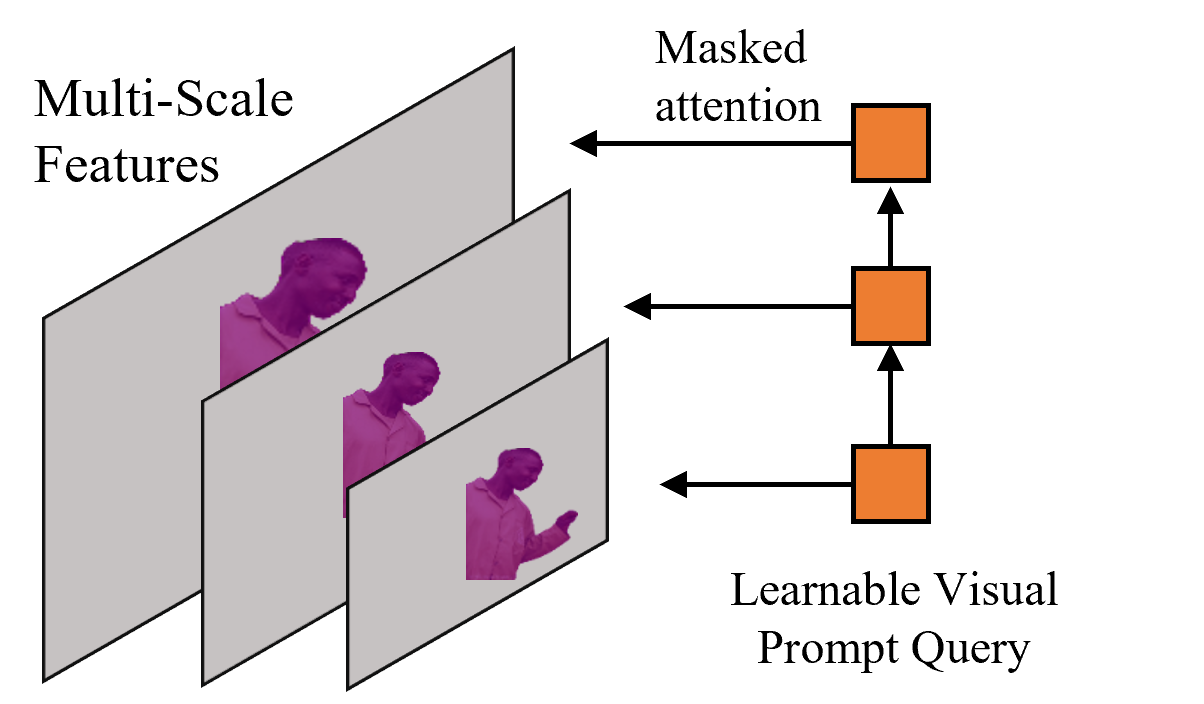}
    % \vspace{-5pt}
    \caption{Prompt encoder to encode visual prompt from reference images. We use three masked cross-attention from the vision encoder small feature map to large feature map.  }
    \label{fig:prompt_enc}
    \vspace{-12pt}
\end{figure}
\begin{figure*}[t!]
    \centering
    \includegraphics[width=\textwidth]{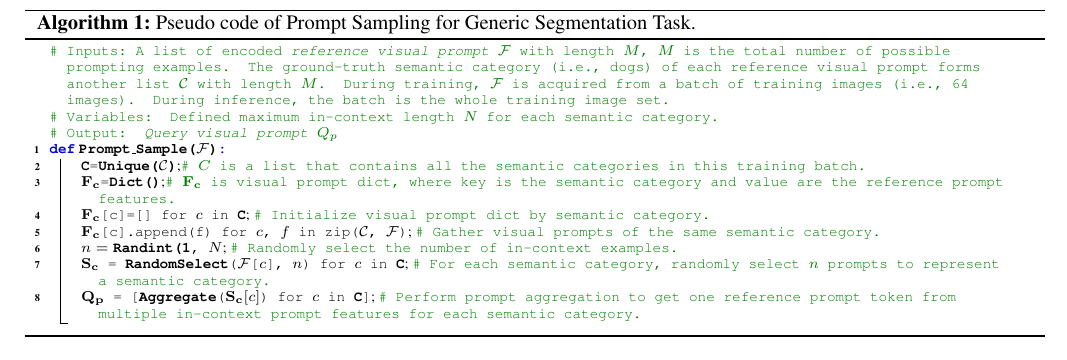}
    \vspace{-5pt}
    \label{fig:algo}
    % \vspace{-12pt}
\end{figure*}
\noindent\textbf{PromptClassifier. }
% \hao{I think we can call this paragraph prompt classifier, ``output fomat" is confusing}
We clarify the definition of the prompt classifier, denoted as $\textbf{PromptClassifier}(\cdot,\cdot)$, for both generic segmentation and referring segmentation tasks here. In the case of generic segmentation tasks like instance and panoptic segmentation, the typical objective is to classify object features $\mathbf{O_s}$ into respective categories. When employing visual prompting for generic segmentation tasks, the distinction lies in the utilization of visual prompt features $\mathbf{O_p}$ as class embeddings. This is illustrated in the following equation:
\begin{equation}
    \mathbf{C_{g}}=\mathbf{g}(\mathbf{O_s})\mathbf{g}(\mathbf{O_p}^T), \mathbf{C_{g}} \in \mathcal{R}^{N_p\times N_s}
\label{EQ:cls1}
\end{equation}
where $N_p$ and $N_s$ are the number of visual prompts and object features. $g$ is the linear projection for generic segmentation task. Each of $N_s$ objects is classified into one of $N_p$ classes.
% In the context of visual prompting, the generic segmentation task, such as instance and panoptic segmentation, aims to identify all objects in a target image that are visually similar to the prompts. 
% In other words, we classify each output mask proposal into visual prompts. 
For visual referring segmentation, the objective differs. Here, each visual prompt is employed to identify the most closely matched instance within the target image. This task can be framed as a classification problem, where each visual prompt is assigned to a specific instance within the target image. It's important to note that during our training phase, the target image and the reference image are identical. The matching score matrix for referring segmentation is structured as follows:
\begin{equation}
\begin{aligned}
    % \mathbf{C_{generic}}&=\mathbf{g}(\mathbf{O_s})\mathbf{g}(\mathbf{O_p}^T), \mathbf{C_{generic}} \in N_q\times N_s \\
    \mathbf{C_{r}}&=\mathbf{h}(\mathbf{O_p})\mathbf{h}(\mathbf{O_s}^T), \mathbf{C_{r}} \in \mathcal{R}^{N_s\times N_q}\\
\end{aligned}
\label{EQ:cls2}
\end{equation}
$h$ is the linear projection for referring segmentation task. Fig.~\ref{fig:framework_all}(b) and (c) provide an illustrative representation of the two tasks. In our implementation, the generic segmentation task involves finding the most suitable visual prompt for each mask proposal, effectively pivoting the loss from a query to all prompts. Conversely, the referring segmentation task focuses on matching a given visual prompt to a specific mask proposal, with the loss pivot transitioning from a prompt to all proposals. As indicated in Equations~\ref{EQ:cls1} and \ref{EQ:cls2}, the $\mathbf{PromptClassifier}$ for both generic and referring segmentation tasks share a similar formulation. Consequently, they can share the entire framework, except for the two distinct linear layers denoted as $\mathbf{g}$ and $\mathbf{h}$.
% However, the goal slightly deviates from the visual referring segmentation which is aimed at locating exactly the same object in a target image, \textit{e.g.,}, video object segmentation~\cite{cheng2021mivos}. The difference of these tasks is also shown in Fig.~\ref{fig:framework_all}(c). In other words, the former task finds the most matched visual prompt for each mask proposal (loss pivot: a query $\Rightarrow$ all prompts), while the latter one in the other way around matches a given visual prompt to a particular mask proposal (loss pivot: a prompt $\Rightarrow$ all proposals)). Observing this difference, the output side of generic and referring segmentation can be formulated as:
% \begin{equation}
% \begin{aligned}
%     % \mathbf{C_{generic}}&=\mathbf{g}(\mathbf{O_s})\mathbf{g}(\mathbf{O_p}^T), \mathbf{C_{generic}} \in N_q\times N_s \\
%     \mathbf{C_{r}}&=\mathbf{h}(\mathbf{O_p})\mathbf{h}(\mathbf{O_s}^T), \mathbf{C_{r}} \in N_s\times N_q\\
% \end{aligned}
% \label{EQ:cls}
% \end{equation}
% % \jianwei{for the above Eqs, we may need to show the dimensions for better interpretation} 
% where $N_p$ and $N_s$ are the number of visual prompts and segmentation queries, and $g$ and $h$ are two linear projection functions to project the output into different task spaces. 
% \jianwei{I changed to g and h because we usually use lowercase variables to denote linear project or so}. 

\subsection{Visual Prompt Formulation}
% 用clip feature不work， 我们提出了一个简单的promp encoder
% 我们用3层 mask cross attention来提取scribble, mask, box的feature作为visual prompt （没有self attention）。三层mask attention从小的feature map再到大的feature map （1/8），attention mask一直保持不会更新。
% \feng{we use a prompt encoder instead of using clip vision encoder for 2 reasons: 1. clip cannot handle some objects and part data in SAM; 2. the feature contain semantics but not appearance similarity for tracking;3. training with clip vision feature is also difficult to converge on generic segmentation (COCO).}
%

The heart part of our \ourmodel{} is the proposed visual prompting mechanism. As shown in Eq.~\ref{EQ:encoder} and Eq.~\ref{EQ:decoder}, we employ two modules to get the final visual prompt:
\begin{itemize}
\vspace{-5pt}
    \item A $\mathbf{PromptEncoder}$ to encode the \textit{reference visual prompt} $\mathcal{F}$ from the reference image features~(followed by a sampling process to get \textit{query visual prompt} $\mathbf{Q_p}$). 
    \vspace{-5pt}
    \item A $\mathbf{Decoder}$ (shared with the segmentation decoder) to decode outputs for the \textit{target visual prompt} $\mathbf{O_p}$ by interacting with the target image features.
    \vspace{-5pt}
\end{itemize}
% \\\noindent A $\mathbf{PromptEncoder}$ to encode the reference visual prompt $\mathbf{Q_p}$ from the reference image features~(followed by a sampling process). 
% \\\noindent A $\mathbf{Decoder}$ (shared with the segmentation decoder) to decode the target visual prompt $\mathbf{O_p}$ by interacting with the target image features.
% \\
This design allows our model to first encode the \textit{reference visual prompt} 
% \jianwei{this term is not well-defined, not clear to me, make sure it is consistent} 
and then adapt the prompt to the target image in a flexible way.
As we attempt to express visual concepts through visual prompts, 
% $\mathbf{PromptEncoder}$ is to some extent similar to open-set feature encoding. visual feature encoding approach in previous image-prompt models
a straightforward way is to employ a pre-trained vision encoder (\textit{e.g.}, CLIP~\cite{clip}) to process the reference images guided by user prompts~\cite{minderer2022simple}. However, it may encounter several challenges: $(i)$ the vision encoder takes cropped images as inputs, which causes substantial domain shift, especially for small objects~\cite{zhong2022regionclip}; $(ii)$ The visual features extracted from CLIP tend to be more semantic and may not meet the demands in VOS tasks. As we will show in our ablation study, employing a CLIP vision encoder to extract visual prompts has a clear inferior generalization ability.

To address these issues, we reuse the vision encoder in our model and develop a simple yet effective prompt encoder. It extracts visual features corresponding to the locations indicated by various forms of visual prompts. To capture visual details of different granularities, we have incorporated multiple layers (default to 3) of the Mask Cross Attention Layer, as shown in Fig.~\ref{fig:prompt_enc}. Each layer takes the image features extracted at different levels (output multi-scale features from the vision encoder, ranging from lower to higher resolutions) as inputs, utilizes the regions defined by the visual inputs as masks, and employs learnable queries to process the features at the corresponding positions to get the visual prompt features.

% In the decoder stage, we further conduct cross-attention operations between visual prompt features and visual tokens to obtain the final visual concept features, which are used as ground-truth visual concepts for computing the contrastive loss with segmentation queries. To ensure that these visual concept queries are not influenced by other information, we have implemented attention masks to prevent segmentation queries and visual prompt features from attending to each other and also ensure that individual visual prompt features are unable to observe and attend to each other.\feng{We may use a small figure for this multi-scale mask attention}

% visual prompt 会过一遍decoder去attend到当前图片的所有visual tokens得到最终用于分类的visual prompt；这里用会用attention mask让segmentation query喝visual prompt query互相不可见，保持独立性； 每个visual prompt彼此也不可见。
% \input{latex/resources/tex/pesudo_code}
% \input{latex/resources/table/main_table}
% \input{latex/resources/table/odinw}
\subsection{Prompt Sampling}
\label{prompt_sample}
% 对于tracking，我们用单张image自己找自己
% 对于detection，我们构建正负样本，基于语义把一个batch里面所有图片的label放在一起，如一共有64 batchsize共50个类别，每个类别一共有1-20个instance，对于每个类别我们sample其中的1-5个instance并 average成为一个token
% \begin{equation}
% \begin{aligned}
%     \langle\mathbf{O_s}, \mathbf{O_p}\rangle &= \mathsf{Dec}\left(\mathbf{Z_i}; \langle\mathbf{Q_s}, \mathbf{Q_p}\rangle\right) \\
%     Q_p^i=Enc_p(P_i|Z_i)\\
%     Q_p=Gather(Q_p^i, ...) \in N\times hid
%     \end{aligned}
%     \label{EQ:DINOv_overall}
% \end{equation}
% \\\textbf{Inference: }
% \input{latex/resources/tex/algo_image}
We introduce two prompt sampling strategies tailored for referring segmentation and generic segmentation, respectively.\\
\noindent\textbf{Referring segmentation.} In the case of referring segmentation, we employ a ``self-referring" approach during training, wherein the reference image is identical to the target image. Here, we sample a prompt from an instance and train the model to refer to the same instance. This approach allows us to leverage extensive segmentation data, such as SA-1B, for training our model effectively.
% As an image in SA-1B can have a large number of masks (i.e., more than 300 masks), we sample a subset of all masks as visual prompts.
Despite being trained on the same instances, our model demonstrates the capability to perform cross-image referring during inference. As illustrated in Fig.~\ref{fig:framework_all}(c), we can change the target images to various different images, enabling the model to effectively engage in cross-image referring tasks.\\
% Therefore, a visual prompt will refer to itself during training and generalize across images as illustrated in Fig.~\ref{fig:framework_all} (c). \\
% \noindent\textbf{} We adopt 
\noindent\textbf{Generic segmentation. } The sampling strategies are slightly different during training and inference:
\begin{itemize}
    \item \textbf{Training}. 
    % For generic segmentation tasks that segment and classify each mask proposal into the correct concepts, 
    In the training process, it is crucial to create both positive and negative visual prompt samples. To achieve this, we generate visual prompts by utilizing a large image training batch.
    % The sampling process begins after we acquire the reference visual prompt features $\mathcal{F}$ for each visual prompt in an image batch.
    As depicted in Algorithm~1, our approach begins by grouping together \textbf{reference visual prompt} $\mathcal{F}$ of the same semantic category across all images within a training batch. For each semantic category, we then randomly select a variable number of in-context examples, ranging from 1 to $N$, and perform an aggregation process to generate \textit{reference visual prompt} tokens $\mathbf{Q_p}$, where each  \textit{reference visual prompt} token corresponds to a specific semantic category. $\mathbf{Q_p}$ is subsequently fed into the decoder, where it interacts with the target image to produce the final \textit{target visual prompt} $\mathbf{O_p}$. Consequently, we attain the same number of target visual prompts to semantic categories. It is noteworthy that a given batch of images may not encompass all semantic categories present in the dataset, resulting in a variable number of semantic categories during the training process.
    \item \textbf{Inference}. During the inference stage, using the COCO dataset as an example, we pre-extract the respective visual prompt features based on mask prompts for all semantic categories established during the training phase. For evaluation purposes, we adopt a random selection approach, where we choose $\mathbf{N}$ (16 by default) features for each semantic category. These selected features act as representative visual prompt features for each category. This practice ensures that our inference stage maintains the same number of categories as in traditional open-set evaluation, effectively preventing any potential information leakage.
\end{itemize}

\begin{table*}
\caption{\textbf{One suit of weights} for generic visual in-context segmentation on multiple datasets. Our model is trained on COCO and SA-1B data. Note: ``$-$" denotes the model does not have number reported or does not have the ability for the specific task.
$\star$ means it is the test set results. $^\dag$ FC-CLIP adopts a frozen CLIP for open-set (text), we prompt the FC-CLIP with CLIP visual features to simulate visual promoting. $\#$ FC-CLIP and ODISE rely on frozen CLIP and Stable Diffusion knowledge. Mask DINO~\cite{li2023mask} is our baseline for comparison.}
\label{tab:generic_seg}
\footnotesize  \setlength{\tabcolsep}{8.0pt}
\centering
\resizebox{0.999\linewidth}{!}{
\begin{tabular}{l|c|c|cccc|cccc|cc} 
% \specialrule{.12em}{.1em}{.1em}
\toprule
\multirow{2}{*}{Method} & \multirow{2}{*}{Semantic Data}& \multirow{2}{*}{Type} & \multicolumn{4}{c|}{{COCO~(in-domain)}} & \multicolumn{4}{c|}{{ADE~(out-domain)}} & \multicolumn{2}{c}{{SegInW~(out-domain)}} \\
 &  & & PQ & mask AP & box AP & mIoU & PQ & mask AP & box AP & mIoU & AP-Average&AP-Median \\ 
\midrule
Mask2Former-T~\cite{cheng2022masked} &COCO& \multirow{7}{*}{Closed-set} &  53.2 & 43.3 & 46.1 & 63.2 & $-$& $-$& $-$& $-$& $-$\\
Mask2Former-B~\cite{cheng2022masked} &COCO&  &  56.4 & 46.3 & 49.5 & 67.1 & $-$& $-$& $-$& $-$& $-$& $-$\\
Mask2Former-L~\cite{cheng2022masked} &COCO&  &  57.8 & {48.6} & 52.1 & 67.4 & $-$& $-$& $-$& $-$& $-$& $-$\\
OneFormer-L~\cite{jain2022oneformer} &COCO&  &  57.9 & {48.9} & $-$ & 67.4 & $-$& $-$& $-$& $-$& $-$& $-$\\
\cellcolor[HTML]{ECF4FF}MaskDINO-L~\cite{li2022mask} &\cellcolor[HTML]{ECF4FF}COCO&  & \cellcolor[HTML]{ECF4FF}58.3 & \cellcolor[HTML]{ECF4FF}{50.6} &\cellcolor[HTML]{ECF4FF} 56.2 & \cellcolor[HTML]{ECF4FF}67.5 & $-$& $-$& $-$& $-$& $-$& $-$\\
Pano/SegFormer-B~\cite{xie2021segformer} &COCO&  &  55.4 & $-$ & $-$ & $-$ & $-$& $-$& $-$& $-$& $-$& $-$\\
kMaX-DeepLab-L~\cite{yu2022k} &COCO&  &48.7 &  {58.1} & $-$ & $-$ & $-$ & $-$& $-$& $-$& $-$& $-$\\ 
\midrule
% GLIPv2-T~\cite{zhang2022glipv2} && \multirow{8}{*}{Text Open-set} &$-$ & 42.0$^\star$ & $-$  \\
GLIPv2-H~\cite{zhang2022glipv2} &COCO+O365+GOLDG+...& \multirow{6}{*}{Text Open-set} &$-$ & 48.9$^\star$ & $-$ & $-$& $-$& $-$& $-$& $-$& $-$& $-$ \\
MaskCLIP~(L)~\cite{ding2022open} &YFCC100M&  & $-$& $-$ & $-$ & $-$ & $-$ & 15.1 & 6.0 & $-$ & 23.7&$-$ \\
\#ODISE-H~\cite{xu2023open}   &COCO~(Stable diffusion))&&45.6&38.4&$-$&52.4& $23.4$ &$13.9$ & $-$ & $28.7$&$-$&$-$\\
\#FC-CLIP-L~\cite{yu2023convolutions}   &COCO~(CLIP)&&54.4&44.6&$-$&63.7& $26.8$ &$16.8$ & $-$ & $34.1$&$-$&$-$\\
{OpenSeed-T}~\cite{zhang2023simple} &COCO+O365&  &{{55.4}} & {47.6} & {52.0} & 64.0& $19.8$ & $14.1$ & $17.0$& $22.9$&$33.9$&21.5 \\
X-Decoder-T~\cite{zou2022generalized} &COCO+CC3M+..&  & 51.4 & 40.5 & 43.6 & 62.8&18.8& 9.8&$-$&25.0&22.7&15.2\\
X-Decoder-L~\cite{zou2022generalized} &COCO+CC3M+..&  & 56.9 & 46.7 & $-$ & 67.5&21.8 &13.1&$-$&29.6&36.1&38.7\\
OpenSeed-L~\cite{zhang2023simple} &COCO+O365&  &  {{59.5}}  & {{53.2}} & {{58.2}}& {{68.6}}& $19.7$ & $15.0$ & $17.7$& $23.4$ &$36.1$&38.7\\
\midrule
FC-CLIP$^\dag$-L~\cite{yu2023convolutions}   &COCO&&$-$&$-$&$-$&$-$& $2.3$ &$4.1$ & $-$ & $7.8$&$-$&$-$\\
SegGPT-L~\cite{wang2023seggpt}&COCO+ADE+VOC+..& \multirow{4}{*}{Visual Prompt} &$43.4$ & $-$ & $-$ & $-$& $-$ & $-$ & $-$& $-$&$-$&$-$ \\
Painter-L~\cite{wang2023images}&COCO+ADE+NYUv2&  &$34.4$ & $-$ & $-$ & $-$& $-$ & $-$ & $-$& $-$&$-$&$-$ \\
\cellcolor[HTML]{FEECE2}\ourmodel{}-T~(Ours)&\cellcolor[HTML]{FEECE2}COCO&  &\cellcolor[HTML]{FEECE2}{$49.0$} & \cellcolor[HTML]{FEECE2}41.5 & \cellcolor[HTML]{FEECE2}$45.2$ & \cellcolor[HTML]{FEECE2}$57.0$& \cellcolor[HTML]{FEECE2}$19.4$ & \cellcolor[HTML]{FEECE2}$11.4$ & \cellcolor[HTML]{FEECE2}$12.8$& \cellcolor[HTML]{FEECE2}$21.9$& \cellcolor[HTML]{FEECE2}$39.5$ & \cellcolor[HTML]{FEECE2}$41.6$ \\
\cellcolor[HTML]{FEECE2}\ourmodel{}-L~(Ours)&\cellcolor[HTML]{FEECE2}COCO&  &\cellcolor[HTML]{FEECE2}$57.7$ & \cellcolor[HTML]{FEECE2}50.4 & \cellcolor[HTML]{FEECE2}$54.2$ & \cellcolor[HTML]{FEECE2}$66.7$ &\cellcolor[HTML]{FEECE2}$23.2$&\cellcolor[HTML]{FEECE2}$15.1$&\cellcolor[HTML]{FEECE2}$14.3$&\cellcolor[HTML]{FEECE2}$25.3$&\cellcolor[HTML]{FEECE2}$40.6$&\cellcolor[HTML]{FEECE2}$44.6$ \\
\bottomrule
% \specialrule{.12em}{.1em}{.1em}
\end{tabular}
}
\end{table*}

\begin{table}[t]
% \vspace{-.4cm}
    \centering
    %     \footnotesize
    %         \renewcommand{\arraystretch}{1.6}
    % \resizebox{\columnwidth}{!}{%
    % \resizebox{\textwidth}{!}{%
    \caption{\textbf{One suit of weights} on ODinW benchmark. Average and median AP across 35 datasets are reported for simplicity.}
    \begin{adjustbox}{width=0.45\textwidth,center}
    % \begin{tabular}{l|c|cccccc}
    \begin{tabular}{l|c|c|ccc|ccc}
        \hline
        Model  & Pretrain Data & Average & Median   \\
        \hline
        % \midrule
        MDETR~\cite{kamath2021mdetr} &GOLDG, REFC &10.7&3.0\\
        GLIP-T~\cite{li2022grounded}    & Object365 &11.4&1.6\\
        
        OpenSeed (T) (Ours)    &Object365, COCO&{14.2}&{3.1}   \\

        OpenSeed (L) (Ours)  &Object365, COCO& {15.2}&{5.0} \\
        \ourmodel{} (T) (Ours)    &COCO, SAM&{14.9}&{5.4}   \\
        \ourmodel{} (L) (Ours)    &COCO, SAM&{15.7}&{4.8}   \\
        \hline
    \end{tabular}
    \end{adjustbox}
    \centering
    % \vspace{0.2cm}

    \label{tab:odinw}
    \vspace{-1.6em}
\end{table}

\subsection{Decoder Query Formulation}

In \ourmodel{}, we designed two types of segmentation queries to address two different tasks as depicted in Fig.~\ref{fig:query_formulation}. For generic segmentation, the query is a number of learnable ones similar to MaskDINO~\cite{li2022mask}. For the visual referring task, we adopt the interactive point query following Semantic-SAM~\cite{li2023semantic}, so that we can exploit the rich granularities in SA-1B~\cite{kirillov2023segment}.
% The main reason is that SA-1B~\cite{kirillov2023segment} data contain rich granularities, which cannot be trivially accommodated by generic queries (around 300). 
% This can be handled by point query proposed in Semantic-SAM~\cite{li2023semantic}, which is also based on MaskDINO. 
Similar to Semantic-SAM, the visual prompts (points or boxes) are both converted into anchor box format, and then the position of each visual prompt will be encoded into position queries. Each position query is duplicated and subsequently combined with content queries of different granularities as the final segmentation queries. For the training on SA-1B, in order to avoid excessive computational overhead on the model, we selectively sample a subset of points contained within this visual concept as positive point queries. Concurrently, we randomly sample a subset of points from the remaining areas to serve as negative points. During the inference stage, we sample the initial point position queries on $20\times20$ uniformly distributed grid as the initial point position for a single frame. 
% \hao{The query type names should be unified with Fig.4}

% 对于coco的detection，我们的segmentation query就是300个learnable的query
% 对于tracking，query采用semantic-sam的query。先简单介绍一下semantic sam的decoder的query形式；
% 训练和inference的tracking query的point的sample机制不一样（train: sample some pos/neg; inference: grid sample 20x20 points）
% \subsection{Query Formulation}
% Interactive query and panoptic query.
\begin{table*}
\centering
\caption{\textbf{Zero-shot} video object segmentation. Without training with video or pairwise image data, our approach is able to do video object segmentation in a zero-shot manner. (\#Concurrent work.) }
% \jianwei{why highlight seem in this table?}
\label{tab:vos}
\resizebox{1.0\linewidth}{!}{
\begin{tabular}{ll|c|c|c|ccccccccccc} 
\toprule
\multirow{2}{*}{Method} & \multirow{2}{*}{Segmentation Data}& \multirow{2}{*}{Type}& \multirow{2}{*}{Refer-Type} & \multirow{2}{*}{\begin{tabular}[c]{@{}c@{}}Zero-\\Shot\end{tabular}}  & \multicolumn{3}{c}{DAVIS17}& \multicolumn{3}{c}{DAVIS16-Interactive}& \multicolumn{5}{c}{YouTube-VOS 2018}\\
& && && JF & J& F& JF & J& F& G& Js & Fs & Ju & Fu\\ 
\midrule
\rowA\multicolumn{16}{l}{\textit{{With Video Data }}} \\
{AGSS}~\cite{lin2019agss}& {VOS+DAVIS}& \multirow{7}{*}{{Video}} & {Mask}& & {67.4} & {64.9} & {69.9} & $-$& $-$& $-$& {71.3} & {71.3} & {65.5} & {75.2} & {73.1}\\
{AGAME}~\cite{johnander2018generative} & {(Synth)VOS+DAVIS}&& {Mask}& & {70.0} & {67.2} & {72.7} & $-$& $-$& $-$& {66.0} & {66.9} & {*}& {61.2} & {*} \\
{SWEM}~\cite{SWEM}& {Image+VOS+DAVIS} && {Mask}& & {84.3} & {81.2} & {87.4} & $-$& $-$& $-$& {82.8} & {82.4} & {86.9} & {77.1} & {85.0}\\
{XMem}~\cite{cheng2022xmem}& {Image+VOS+DAVIS} && {Mask}& & $-$ & $-$ &$-$& $-$& $-$& $-$& {86.1} & {85.1} & {89.8} & {80.3} & {89.2}\\
{SiamMask}~\cite{wang2019fast}& {COCO+VOS}&& {Box} & & {*}& {54.3} & {58.5} & {69.8} & {71.7} & {67.8} & {*}& {60.2} & {58.2} & {45.1} & {47.7}\\
{MiVOS}~\cite{cheng2021mivos} & {BL30K+VOS+DAVIS} && {Mask} & & {84.5} & {81.7} & {87.4} & {91.0} & {89.6} & {92.4} & {82.6} & {81.1} & {85.6} & {77.7} & {86.2}\\
{ReferFormer-B}~\cite{wu2022referformer} & {RefCOCO(+/g)+VOS+DAVIS}&& {Text}& & {61.1} & {58.1} & {64.1} & $-$& $-$& $-$& {*}& {*}& {*}& {*}& {*} \\ 
\hline
% {TAM-L}~\cite{yang2023track} & {XMem+SAM}& \multirow{4}{*}{{Generalist}}& {Multiple Points} & & & $-$& $-$& $-$& {88.4} & {87.5} & {89.4} & $-$& $-$& $-$& $-$& $-$ \\
{UNINEXT-T}~\cite{yan2023universal} & {Image+Video }&\multirow{3}{*}{{Generalist}}& {Mask}& & {74.5} & {71.3} & {77.6} & {$-$}& {$-$}& {$-$}& {77.0} & {76.8} & {81.0} & {70.8} & {79.4}\\
{UNINEXT-L}~\cite{yan2023universal} & {Image+Video}&& {Mask}& & {77.2} & {73.2} & {81.2} & {$-$}& {$-$}& $-$& {78.1} & {79.1} & {83.5} & {71.0} & {78.9}\\
{UNINEXT-L}~\cite{yan2023universal} & {Image+Video}&& {Text}& & {66.7} & {62.3} & {71.1} & {$-$}& {$-$}& {$-$}& {*}& {*}& {*}& {*}& {*} \\ 
\midrule
\rowA\multicolumn{16}{l}{\textit{{Without Video Data }}} \\
Painter-L~\cite{wang2023images} & COCO+ADE+NYUv2 & \multirow{7}{*}{Generalist}& Mask& \cmark& 34.6 & 28.5 & 40.8 & $-$ & $-$ &$-$& 24.1 & 27.6 & 35.8 & 14.3 & 18.7\\
SegGPT-L~\cite{wang2023seggpt}& COCO+ADE+VOC+... && Mask& \cmark& 75.6 & 72.5 & 78.6 & $-$ &$-$& $-$ & 74.7 & 75.1 & 80.2 & 67.4 & 75.9\\
PerSAM-L~\cite{zhang2023personalize}& SAM+DAVIS && Mask& \xmark& 60.3 & 56.6 & 63.9 &$-$& $-$ & $-$ & * &*& * & * & *\\
SEEM-T~\cite{zou2023segment}& && & \cmark& {60.4} & {57.6} & {63.3} & {62.7} & {58.9} & {66.4} & {51.4} & {55.6} & {44.1} & {59.2} & {46.9}\\
SEEM-L~\cite{zou2023segment}& \multirow{-2}{*}{COCO+LVIS} && \multirow{-2}{*}{Mask} & \cmark& {58.9} & {55.0} & {62.8} & {62.2} & {58.3} & {66.0} & {50.0} & {57.2} & {38.2} & {61.3} & {43.3}\\

\cellcolor[HTML]{FEECE2}\ourmodel{}-T~(Ours)& \multirow{2}{*}{COCO+SAM} &&{Mask}&\cellcolor[HTML]{FEECE2}\cmark&\cellcolor[HTML]{FEECE2}73.3&\cellcolor[HTML]{FEECE2}71.0&\cellcolor[HTML]{FEECE2}75.7&\cellcolor[HTML]{FEECE2}77.0&\cellcolor[HTML]{FEECE2}72.9&\cellcolor[HTML]{FEECE2}81.2&\cellcolor[HTML]{FEECE2}60.9&\cellcolor[HTML]{FEECE2}65.3&\cellcolor[HTML]{FEECE2}70.0&\cellcolor[HTML]{FEECE2}52.3&\cellcolor[HTML]{FEECE2}57.9 \\
\cellcolor[HTML]{FEECE2}\ourmodel{}-L~(Ours)&  &&&\cellcolor[HTML]{FEECE2}\cmark&\cellcolor[HTML]{FEECE2}72.3&\cellcolor[HTML]{FEECE2}69.8&\cellcolor[HTML]{FEECE2}74.8&\cellcolor[HTML]{FEECE2}75.4&\cellcolor[HTML]{FEECE2}71.3&\cellcolor[HTML]{FEECE2}79.4&\cellcolor[HTML]{FEECE2}59.6&\cellcolor[HTML]{FEECE2}61.7&\cellcolor[HTML]{FEECE2}65.7&\cellcolor[HTML]{FEECE2}52.3&\cellcolor[HTML]{FEECE2}58.8 \\
\hline
\end{tabular}
}
\end{table*}
\section{Experiments}
\subsection{Setup}
\noindent\textbf{Dataset and Settings. }In our experiments, we jointly train on two types of data: segmentation data with semantic labels and segmentation data with only pixel annotations (SA-1B~\cite{kirillov2023segment}. For semantically-labeled data, we use COCO2017~\cite{lin2014microsoft} panoptic segmentation dataset with around 110K images. For SA-1B, we employ a 20\%  portion subset with around 2M images. We evaluate our model on a wide range of tasks and datasets with only visual prompts, including: 1. Open-set panoptic segmentation on COCO2017~\cite{lin2014microsoft} and ADE20K~\cite{zhou2017scene}; 2. Segmentation in the wild (SegInW)~\cite{zou2022generalized} which includes 25 instance segmentation datasets; 3. Object detection in the wild (ODinW)~\cite{li2022grounded} that encompasses over 35 datasets; 4. Zero-shot Video object segmentation (VOS) on DAVIS2017~\cite{pont20172017}, DAVIS2016-Interactive~\cite{pont20172017}, and Youtube-VOS 2018~\cite{xu2018youtube}.
\\
\noindent
\textbf{Implementation Details. } We provide implementation details in the Appendix.
% Our model framework is mainly based on Mask DINO~\cite{li2023semantic}, which is a unified framework for detection and segmentation. \ourmodel{} is a general encoder-decoder architecture composed of a vision encoder. We use Swin-T/L~\cite{liu2021swin} as the vision encoder. As our decoder supports both generic query and point query, we adopt 300 latent generic queries following Mask DINO~\cite{li2022mask} and six level queries for each input point following Semantic-SAM~\cite{li2023semantic}. Specially, when using point query, we sample 10 foreground and 40 background points during training and employ grid sample for 20$\times$20 points during inference. For inference on general segmentation and detection tasks, we use 16 in-context examples for each category by default. For VOS inference, we average eight previous frame predictions as the reference to segment the current frame. 
\\
\noindent
\textbf{Evaluation Metrics. }For all segmentation and detection tasks, we use standard evaluation metrics: PQ (Panoptic Quality) for panoptic segmentation, AP (Average Precision) for instance segmentation (mask AP) and detection (box AP), and mIoU (mean Intersection over Union) for semantic segmentation. For VOS tasks, we follow previous semi-supervised models to use region similarity J and contour accuracy F. We also adopt the averaged score J\&F as the metric for DAVIS2017 and averaged overall score G for Youtube-VOS 2018. Note that Youtube-VOS 2018 also reports J and F for seen and unseen splits.
\subsection{Generic Segmentation and Detection}
We evaluate our visual prompt based generic segmentation performance in Table~\ref{tab:generic_seg}.
\\\noindent\textbf{In-domain Segmentation on COCO. } Compared to other models trained for visual prompts, we achieve significantly better results. For example, we surpass SegGPT~\cite{wang2023seggpt} and Painter~\cite{wang2023images} by $14.3$ PQ and $25.5$ PQ.   In addition, With just a few visual in-context prompts for each category, our model achieves comparable results with previous close-set or open-set models on COCO. For example, the panoptic segmentation performance gap between \ourmodel{} and our baseline Mask DINO is only $0.6$ PQ ($57.7$ PQ vs $58.3$ PQ).
\\ 
\noindent 
\textbf{Out-domain open-set segmentation on ADE20K. }After training with visual prompt on COCO and SAM, we do zero-shot evaluation on ADE20K to validate its open-set segmentation capability when seeing novel visual concepts. To our best knowledge, it is the first time to use visual prompt for open-set segmentation. Compared with previous text-prompted open-set models, we achieve comparable or better performance with only COCO semantic data and no semantic knowledge from large pre-trained models. Especially, compared with our baseline OpenSeed, we achieve better performance with much fewer data. Note that FC-CLIP~\cite{yu2023convolutions} employs a frozen CLIP to do text-based open-set segmentation. As the text and visual features are aligned in CLIP, we also attempt to prompt a pre-trained FC-CLIP with visual features from CLIP to test its open-set ability with visual prompts. However, its visual prompting performance largely lags behind its text-prompted results. Therefore, it is non-trivial to transfer a multi-modal text-based open-set model to do visual-prompted recognition well. The results indicate that visual prompts can generalize well to new concepts.
\\ 
\noindent 
\textbf{Segmentation and detection in the wild. }We also validate the generalization capability of visual prompting on some diversified and domain-specific datasets including SegInW and ODinW, which in total encompass more than 60 datasets. These datasets contain many real-scenario or rare categories. As these datasets all focus on instance-level segmentation, we report the average and median AP (AP-Average and AP-Median) over all datasets. We first evaluate the Segmentation in the Wild (SegInW) benchmark, which consists of 25 datasets. With visual prompting, \ourmodel{} achieves a significant performance improvement over our baseline OpenSeed. For example, Our best AP-Average exceeds OpenSeed by 4.5 AP. We further evaluate Object Detection in the Wild (ODinW), which is composed of 35 datasets with bounding box annotations. As shown in Table~\ref{tab:odinw}, though we only employ much fewer semantically labeled data, we achieve better performance compared with previous models under similar settings.
\subsection{Video Object Segmentation}
Video object segmentation (VOS) aims to segment an interested object in a video by giving text or visual clues. Our model focuses on the semi-supervised setting, which segments a particular object throughout a video by giving visual clues in the first frame. In \ourmodel{}, the visual prompt originates from one single image (generic/referring segmentation) or other images in one batch~(generic segmentation). Therefore, our model has learned to prompt with visual features from other images. Therefore, \ourmodel{} is able to do video object segmentation (VOS) by replacing current frame visual prompt features with previous frames. For more accurate tracking, we also store the visual features of the predicted mask in previous frames. These features, denoted as memory visual prompts, will be averaged together with the first frame's given prompt to construct the visual prompt of the current frame. Details of the memory visual prompt and ablations are in the Appendix. By default, the memory length is set to 8.  In Table~\ref{tab:vos}, we conduct (interactive) video object segmentation evaluation on DAVIS17, DAVIS2016-Interactive, and Youtube-VOS 2018. The results of DAVIS2017 and Youtube-VOS 2018 indicate our model achieves better performance than SEEM and PerSAM. In addition, \ourmodel{} can also do interactive VOS, and our performance on DAVIS16-Interactive achieves significantly better performance compared with models not using video data for training.
\subsection{Ablation}
\begin{figure}[t!]
    \centering
    \includegraphics[width=0.47\textwidth]{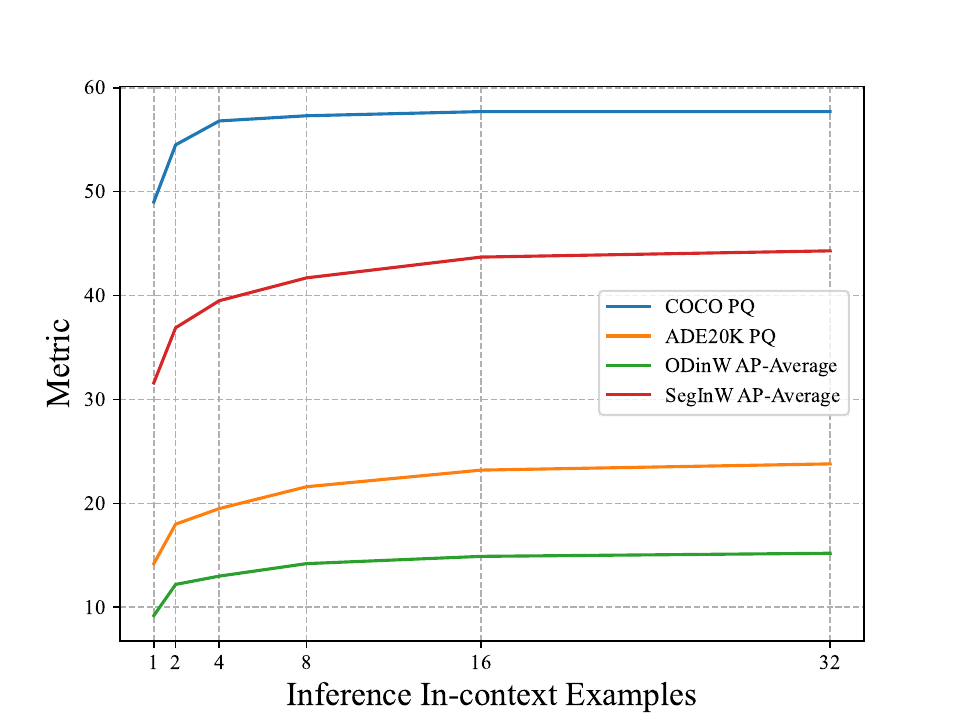}
    \vspace{-1pt}
    \caption{\ourmodel{} query formulation of different tasks.  }
    \label{fig:inference_example_num}
    \vspace{-12pt}
\end{figure}

\noindent \textbf{Effectiveness of Query Formulation. } In Table~\ref{tab:ablation:query_remove}, we ablate the effectiveness of using different query formulations for different tasks. The results indicate our double query formulation outperforms using only one type of query.
\\
\noindent \textbf{Effectiveness of Visual Prompt Formulation. } In Table~\ref{tab:ablation:prompt_encoder}, we attempt to use a pre-trained CLIP vision encoder to encode the features of the visual prompt by cropping the prompted region into images for CLIP to process. As CLIP features contain rich semantics with few appearance features, which could not apply to referring segmentation tasks. Therefore, we ablate on generic segmentation tasks and find that the final model could not generalize well on open-set datasets like ADE. This result verifies our hypothesis that CLIP vision features could not generalize well on in-context visual prompting.
\\
\noindent \textbf{Effectiveness of  Unifying Tasks and Data. }We unify visual generic segmentation and visual referring segmentation to use both semantically labeled data (COCO) and data with only segmentation annotations (SA-1B). In Table~\ref{tab:ablation:data_unification}, the results indicate that employing both datasets improves each individual task.
\\
\noindent \textbf{Training batch size for generic segmentation. } In Table~\ref{tab:ablation:train_batch}, the results show that increasing training batch size consistently improves the generic segmentation performance. The reason for this phenomenon is that a larger batch size helps to sample more positive and negative visual in-context examples across different images, which better matches the inference setting with random visual examples.
\\
\noindent \textbf{Inference In-Context Examples. } In Fig.~\ref{fig:inference_example_num}, we ablate the impact of using different in-context lengths. Increasing the in-context example exhibits diminishing returns, especially when the number of examples is more than eight.

\begin{table}
\centering
% \tablestyle{3pt}{0.55}
% \setlength{\extrarowheight}{0pt}
\addtolength{\extrarowheight}{\aboverulesep}
% \addtolength{\extrarowheight}{\belowrulesep}
% \setlength{\aboverulesep}{0pt}
% \setlength{\belowrulesep}{0pt}
\caption{\textbf{Ablation} of using difference queries to do both in-context reference and generic segmentation. By default, we use both generic query and interactive query. We remove one type of query at a time to ablate their effectiveness.}
\label{tab:task_transfer}
\begin{adjustbox}{width=0.49\textwidth,center}
\begin{tabular}{l|cccc|ccc} 
\toprule
\multirow{2}{*}{Method}                        
&\multicolumn{4}{c|}{COCO}   & \multicolumn{3}{c}{DAVIS17} \\
& PQ  & mask AP  & box AP & {mIoU}    & JF & J & F \\ 
\midrule
\ourmodel{}-SwinT  & 49.6&42.7&47.0&58.0&73.3&71.0&75.7 \\
only point query  &  45.2 & \textbf{31.0}\fontsize{8.0pt}{\baselineskip}\selectfont{(\textbf{11.7})} &\textbf{34.7}\fontsize{8.0pt}{\baselineskip}\selectfont{(\textbf{-12.3})}&52.7&71.4&68.8&74.0 \\
only generic query &  46.2 & \textbf{38.3}\fontsize{8.0pt}{\baselineskip}\selectfont{(\textbf{-4.4})} &\textbf{41.5}\fontsize{8.0pt}{\baselineskip}\selectfont{(\textbf{-6.0})}&53.3&   68.9&66.5&71.3                                          \\
\bottomrule
\end{tabular}
\end{adjustbox}
\label{tab:ablation:query_remove}
\vspace{-.1cm}
\end{table}

\begin{table}
\centering
% \tablestyle{3pt}{0.55}
% \setlength{\extrarowheight}{0pt}
\addtolength{\extrarowheight}{\aboverulesep}
% \addtolength{\extrarowheight}{\belowrulesep}
% \setlength{\aboverulesep}{0pt}
% \setlength{\belowrulesep}{0pt}
\caption{\textbf{Ablation} of using different ways to encode the visual prompt on our Swin-T model. Under the same setting, we change our prompt encoding method and use a pre-trained CLIP to crop and encode the prompted objects in the image.}
\vspace{-3pt}
\label{tab:task_transfer}
\begin{adjustbox}{width=0.49\textwidth,center}
\begin{tabular}{l|cccc|cccc} 
\toprule
\multirow{2}{*}{\begin{tabular}[c]{@{}c@{}}Prompt\\ Encoding \end{tabular}}                        
&\multicolumn{4}{c|}{COCO~(in-domain)}   & \multicolumn{4}{c}{ADE~(out-domain)} \\
& PQ  & mask AP  & box AP & {mIoU} & PQ  & mask AP  & box AP & {mIoU}\\ 
\midrule
Ours  & 49.6&42.7&47.0&58.0&19.4&11.4&12.8&21.9 \\
CLIP & 48.5 &40.7&43.5 &54.9  &12.6&1.4&1.3&13.3            \\
\bottomrule
\end{tabular}
\end{adjustbox}
\label{tab:ablation:prompt_encoder}
\vspace{-0.3cm}
\end{table}

\begin{table}
\centering
% \tablestyle{3pt}{0.55}
% \setlength{\extrarowheight}{0pt}
\addtolength{\extrarowheight}{\aboverulesep}
% \addtolength{\extrarowheight}{\belowrulesep}
% \setlength{\aboverulesep}{0pt}
% \setlength{\belowrulesep}{0pt}
\caption{\textbf{Ablation} of the effectiveness of unifying tasks and data.}
\label{tab:task_transfer}
\begin{adjustbox}{width=0.49\textwidth,center}
\begin{tabular}{lc|cccc|ccc} 
\toprule
\multirow{2}{*}{Method}  &\multirow{2}{*}{Data}                      
&\multicolumn{4}{c|}{COCO}   & \multicolumn{3}{c}{DAVIS17} \\
&& PQ  & mask AP  & box AP & {mIoU}    & JF & J & F \\ 
\midrule
\ourmodel{}-SwinT  &COCO, SAM& 49.6&42.7&47.0&58.0&73.3&71.0&75.7 \\
\ourmodel{}-SwinT  & COCO& 48.9&41.7&45.9&57.1&63.3&60.8&65.7 \\
\ourmodel{}-SwinT & SAM&N/A & $-$&$-$&$-$&   68.4&66.0&70.8                                       \\
\bottomrule
\end{tabular}
\end{adjustbox}
\label{tab:ablation:data_unification}
\vspace{-0.3cm}
\end{table}

\begin{table}
\centering
% \tablestyle{3pt}{0.55}
% \setlength{\extrarowheight}{0pt}
\addtolength{\extrarowheight}{\aboverulesep}
% \addtolength{\extrarowheight}{\belowrulesep}
% \setlength{\aboverulesep}{0pt}
% \setlength{\belowrulesep}{0pt}
\caption{\textbf{Ablation} of image batchsize sampling in training.}
\label{tab:task_transfer}
\begin{adjustbox}{width=0.49\textwidth,center}
\begin{tabular}{l|c|cccc} 
\toprule
\multirow{2}{*}{Method}  &\multirow{2}{*}{\begin{tabular}[c]{@{}c@{}}\#Batchsize for\\ Prompt Sampling \end{tabular}}                      
&\multicolumn{4}{c}{COCO}    \\
&& PQ  & mask AP  & box AP & {mIoU}     \\ 
\midrule
\ourmodel{}-SwinT  &1& 28.9&23.2&25.3&33.7 \\
\ourmodel{}-SwinT  & 4& 45.1&37.0&40.4&50.6 \\
\ourmodel{}-SwinT & 8&$47.3$ & $39.2$&$43.1$& 53.1 \\
\ourmodel{}-SwinT & 32&$47.8$ & $40.3$&$44.1$& 56.2 \\
\ourmodel{}-SwinT & 64&$49.0$ & $45.2$&$41.5$& 57.0 \\
\bottomrule
\end{tabular}
\end{adjustbox}
\label{tab:ablation:train_batch}
\vspace{-0.3cm}
\end{table}
\begin{figure*}[t!]
    \centering

    \includegraphics[width=\textwidth]{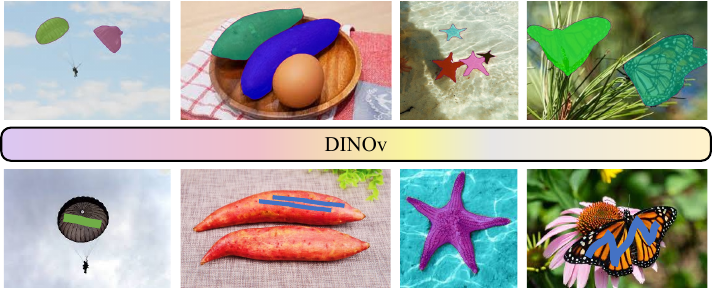}
    \vspace{-15pt}
    \caption{\ourmodel{} can do open-set segmentation by giving visual prompts.
    % (c) An illustration of losses for visual generic segmentation and visual referring segmentation. 
    }    
    % \hao{The caption is too short. Should explain each sub-figure. For (c), I understand each collume denotes a visual prompt, but the meaning of each row is not very clear.}
    \label{fig:framework_all}
    % \vspace{-12pt}
    \vspace{-5pt}
\end{figure*}
\section{Related Works}
\subsection{Visual Perception Through Text Prompt}
Innovations in open-vocabulary object detection \cite{liu2023grounding, gu2021open, zang2022open, zhong2022regionclip, zhang2023simple, minderer2022simple, li2022grounded, kamath2021mdetr} and open-vocabulary segmentation~\cite{zhang2023simple, lan2021discobox, huynh2022open, ghiasi2022scaling, rao2022denseclip, xu2023open}, have shown great potential in generic visual perception, by leveraging large pre-trained vision-language models like CLIP \cite{radford2021learning} and ALIGN \cite{jia2021scaling}. These approaches demonstrate significant strides in zero-shot and few-shot performance, adapting to a variety of visual contexts through text prompts. However, the reliance on text alone introduces limitations due to linguistic ambiguity and the potential mismatch between textual descriptions and complex visual scenes \cite{xu2023multi}. This highlights the ongoing need to refine the integration of visual inputs for more accurate and comprehensive image perception.

\subsection{Visual Perception Through Image Example}
Building upon the foundations set by text-based visual perception methodologies, the field has seen a notable shift towards incorporating image examples to enhance accuracy and context sensitivity. OV-DETR \cite{zang2022open} extends its open-vocabulary object detection capability beyond text, by utilizing both the image encoder and text encoder from CLIP \cite{radford2021learning}, allowing for object detection guided by visual examples. Similarly, OWL-ViT \cite{minderer2022simple} leverages large-scale image text examples in its contrastive pre-training phase, and propose to adopt image example for one-shot image-conditioned object detection. MQ-Det \cite{xu2023multi} utilizes image examples to enhance text descriptions for better open-vocabulary object detection performance. These methods typically adopt the image encoder in CLIP to extract visual features from given image examples for a more accurate perception of objects and scenes, and demonstrate that visual examples can bridge the gap between textual ambiguity and the complex nature of visual perception. 

\subsection{Visual Perception Through Visual Prompt}
Different from image example-based methods that take an image as input, which are then processed by multi-modal encoder like CLIP \cite{radford2021learning}, visual prompt-based methods typically use visual instructions (e.g. box, point, mask, scribble, and refereed regions of another image) to guide a model for a specific visual task. SAM \cite{kirillov2023segment}, for instance, introduces a promptable model for interactive image segmentation, fostering research in computer vision foundation models. It is followed by some works that adapt SAM for visual prompting through personalized examples~\cite{zhang2023personalize}. SEEM \cite{zou2023segment} stands out as an interactive and versatile model for segmenting objects, accommodating various types of prompts, and is semantic-aware compared to SAM. Semantic-SAM \cite{li2023semantic} excels in semantic awareness and recognizing granularity, 
 and is capable of various segmentation tasks including panoptic and part segmentation. Painter~\cite{wang2023images} and SegGPT~\cite{wang2023seggpt} take a generalist approach, coping with various segmentation tasks by formulating segmentation as an in-context coloring problem. Our work resembles them with the same goal while presenting a new visual prompting mechanism to support all types of segmentation tasks.
 
\section{Conclusion}
We present \ourmodel{}, a unified framework for in-context visual prompting to accommodate both referring segmentation and generic segmentation tasks. To effectively formulate in-context visual prompts, we designed a simple prompt encoder to encoder reference visual prompts from the reference image
and adopted a shared decoder to decode the final target visual prompts from the target image. We also formulate generic latent queries and point
queries to align different tasks and data. The experimental results indicate that \ourmodel{} demonstrates impressive referring and generic segmentation capabilities to refer and detect with in-context visual prompting. Notably, \ourmodel{} delivers competitive performance compared to close-set segmentation on in-domain
datasets and show promising results on many open-set
segmentation benchmarks. We hope our early exploration of visual in-context prompting could inspire the community.
\\
\noindent\textbf{Limitations. }We employ limited semantically labeled data (COCO), which can be scaled up for better performance and extended to text prompts for multi-modal understanding. 
% . In addition, \ourmodel{} focuses on visual prompting without explicitly training on text prompting, which could be jointly trained in the future.

%%%%%%%%% REFERENCES
{\small
\bibliographystyle{ieee_fullname}
\bibliography{egbib}
}
\appendix
\section{Implementation Details}
% \textbf{Implementation Details. } 
Our model framework is mainly based on Mask DINO~\cite{li2023semantic}, which is a unified framework for detection and segmentation. \ourmodel{} is a general encoder-decoder architecture composed of a vision encoder. We use Swin-T/L~\cite{liu2021swin} as the vision encoder. As our decoder supports both generic query and point query, we adopt 300 latent generic queries following Mask DINO~\cite{li2022mask} and six level queries for each input point following Semantic-SAM~\cite{li2023semantic}. Especially, when using point query, we sample 10 foreground and 40 background points during training and employ grid sample for 20$\times$20 points during inference. For inference on general segmentation and detection tasks, we use 16 in-context examples for each category by default. For VOS inference, we average eight previous frame predictions as the reference to segment the current frame. 
\\
\section{Video Object Segmentation Inference}
Video object segmentation (VOS) aims to segment an interested object in a video by giving text or visual clues. Our model focuses on the semi-supervised setting, which segments a particular object throughout a video by giving visual clues in the first frame. When doing VOS, an intuitive way is to first extract reference visual prompt features from the first frame image and the corresponding visual prompts with our prompt encoder. When processing each frame in a video, we are able to utilize reference visual prompt features in the first frame as in the current frame. 

In \ourmodel{}, as we train with visual in-context prompting with multiple examples for generic segmentation, we can also apply this strategy to VOS for better performance. More concretely, we also compute and store the reference visual features of the predicted mask in previous frames. These features, denoted as memory reference visual prompts, will be averaged together with the first frame's given prompt to construct the visual prompt of the current frame. We employ a priority queue to manage the memory. For simplicity, the priority score of each prompt is positively correlated to the frame number, which indicates that we only store the memory prompts that are near the current frame in time sequence.  By default, the memory length is set to 8. In Tab.~\ref{tab:ablation:memory_length}, we show the influence of using different number of memory length.

\begin{table}
\centering
% \tablestyle{3pt}{0.55}
% \setlength{\extrarowheight}{0pt}
\addtolength{\extrarowheight}{\aboverulesep}
% \addtolength{\extrarowheight}{\belowrulesep}
% \setlength{\aboverulesep}{0pt}
% \setlength{\belowrulesep}{0pt}
\caption{\textbf{Ablation} of Inference Memory Length on DAVIS2017 with a SwinL backbone.}
\label{tab:task_transfer}
\begin{adjustbox}{width=0.45\textwidth,center}
\begin{tabular}{lc|ccc} 
\toprule
\multirow{2}{*}{Method}  &\multirow{2}{*}{Memory Length}                      
&\multicolumn{3}{c}{DAVIS2017}    \\
&& J\& F  &J  & F    \\ 
\hline
\ourmodel{}-SwinT  &1& 62.1&58.7&65.4 \\
\ourmodel{}-SwinT  & 2& 69.6&66.7&72.6 \\
\ourmodel{}-SwinT  & 4& 71.5&68.7&74.3 \\
\ourmodel{}-SwinT & 8&$72.3$ & $69.8$&$74.8$         \\
\ourmodel{}-SwinT & 16&$68.0$ & $65.4$&$70.7$         \\
% \ourmodel{}-SwinT  & 2& 74.0&71.8&76.3 \\
% \ourmodel{}-SwinT  & 4& 75.5&73.0&78.3 \\
% \ourmodel{}-SwinT & 8&$73.3$ & $71.0$&$75.7$                                   \\

\bottomrule
\end{tabular}
\end{adjustbox}
\label{tab:ablation:memory_length}
\vspace{-0.3cm}
\end{table}
\end{document}